\documentclass[a4paper, 10pt, conference]{ieeeconf}

\usepackage[pdftex]{graphicx}
\graphicspath{{images/}}
\DeclareGraphicsExtensions{.png}

\usepackage{amssymb}
\usepackage[cmex10]{amsmath}

\usepackage[noend]{algpseudocode}
\usepackage{hyperref}

\usepackage{multirow}

\begin{document}

\title{\LARGE \bf
Strengthening the Case for a Bayesian Approach to Car-following Model Calibration and Validation using Probabilistic Programming\\[1.75ex]
\normalfont\large
Franklin Abodo\authorrefmark{1}, Andrew Berthaume\authorrefmark{2}, Stephen Zitzow-Childs\authorrefmark{2} and Leonardo Bobadilla\authorrefmark{1}
\\[-0.75ex]
}

\author{
\authorblockA{
\authorrefmark{1}School of Computing and Information Sciences\\
College of Engineering and Computing\\
Florida International University\\
Miami, FL 33199, USA\\
Email: fabod001@fiu.edu, bobadilla@cs.fiu.edu
}
\and
\authorblockA{
\authorrefmark{2}Volpe National Transportation Systems Center\\
Office of Research and Technology\\
U.S. Department of Transportation\\
Cambridge, MA 02142, USA\\
Email: \{andrew.berthaume, s.zitzow-childs\}@dot.gov
}
}

\maketitle

\begin{abstract}

 Compute and memory constraints have historically prevented traffic simulation software users from fully utilizing the predictive models underlying them. When calibrating car-following models, particularly, accommodations have included 1) using sensitivity analysis to limit the number of parameters to be calibrated, and 2) identifying only one set of parameter values using data collected from multiple car-following instances across multiple drivers. Shortcuts are further motivated by insufficient data set sizes, for which a driver may have too few instances to fully account for the variation in their driving behavior. In this paper, we demonstrate that recent technological advances can enable transportation researchers and engineers to overcome these constraints and produce calibration results that 1) outperform industry standard approaches, and 2) allow for a unique set of parameters to be estimated for each driver in a data set, even given a small amount of data. We propose a novel calibration procedure for car-following models based on Bayesian machine learning and probabilistic programming, and apply it to real-world data from a naturalistic driving study. We also discuss how this combination of mathematical and software tools can offer additional benefits such as more informative model validation and the incorporation of true-to-data uncertainty into simulation traces.
\end{abstract}

\IEEEpeerreviewmaketitle

\section{Introduction}
Traffic simulation software packages are widely used in transportation engineering to estimate the impacts of potential changes to a roadway network and forecast system performance under future scenarios. These packages are underpinned by math- and physics-based models, which are designed to describe behavior at an aggregate (macroscopic) level or at the level of individual drivers (microscopic). Within the microscopic realm, driver behavior is decomposed into sub-models which individually handle lane-changing, car-following, route choice, and so on. This work focuses specifically on car-following models (CFMs), which estimate the acceleration and deceleration behavior of individual vehicles with respect to their driving environment. Critical to the accurate performance of simulation models is the calibration process, during which unobserved model parameters have their values estimated using field measurements. Due to known variations in driver behavior that exist across regions and between driving conditions (such as road weather conditions), every transportation network model must be re-calibrated to accurately reflect the in situ driver behavior prior to forecasting future conditions. A variety of statistical and machine learning tools have been employed to estimate the values of a CFM’s latent variables in prior work. CFMs are typically calibrated using optimization techniques, with the genetic algorithm being a popular approach \cite{ear}, \cite{hammit}. A less frequently used method based on Bayesian inference was shown in \cite{rahman} to outperform one particular deterministic optimization method applied to a real-world data set. However, the application of Bayesian methods to the calibration (and validation) of CFMs remains not thoroughly explored in the literature. As recently as fifteen years ago, the use of Markov Chain Monte Carlo (MCMC) methods for calibration of traffic simulator parameters was considered a difficult challenge that required expert knowledge, mainly when applied at the scale of the large parameter spaces and large field data sets often under consideration \cite{higdon}. In this paper, we revisit the problem of CFM calibration in the context of Bayesian machine learning, demonstrating how the advent of probabilistic programming languages (PPLs) has enabled the convenient construction of probabilistic variations of CFMs that yield more effective calibration results by exploiting the hierarchical structure of field data. This result is of particular interest to researchers and engineers concerned with agent-based modeling and simulation \cite{ear}, in which the driving behavior and travel related decision making of individuals are studied and used in downstream analyses. We also discuss how Bayesian approaches to model criticism allow for more stringent and informative validation of CFMs. High-resolution data from a naturalistic driving study serve as the basis of these demonstrations. The statistical models and calibration procedure are implemented using the distributed computing framework TensorFlow Probability (TFP) and the PPL Edward2 \cite{tran}. The source code is publicly available for other researchers to compare results using different data sets. 

\section{Related Work}

In prior work on calibration for traffic simulators, Bayesian methods have been advocated because of their ability to capture uncertainty in estimated parameter values stemming from: 1) inconsistency between the model and the natural phenomena it is meant to capture; 2) errors introduced by the calibration process itself (including the choices of measure of performance and measure of error (MoE)); and 3) noise or errors in the data collection process \cite{bayarri}, \cite{molina}. In \cite{bayarri} and \cite{molina}, Markov Chain Monte Carlo methods are used to analyze the error in human measurement of turn counts and roadway entry counts, and to estimate other parameters of the CORSIM simulator. In \cite{molina}, the influence of sampling from parameter distributions when generating simulation traces rather than fixing parameter values is additionally evaluated. \cite{zhong} modeled the Intelligent Driver Model using Gaussian random variables as the parameters to perform a probabilistic sensitivity analysis based on the Kullback-Liebler dissimilarity measure in order to limit the number of parameters requiring value estimation to those yielding the greatest performance improvement relative to default parameter values. \cite{rahman} advocated for the use of a Bayesian approach to calibration by directly comparing an MCMC-based calibration method with a deterministic optimization method, using one synthetic data set to show that the Bayesian method could recover known parameter values and one real-world data set as a case study. One common characteristic of these prior works is a lack of transparency into the calibration procedures that would allow their results to be reproduced using the same data (assuming public availability) or contested using data from a different distribution collected under different scenarios. 

In this work, we seek to strengthen the case for Bayesian methods. First, we use a hierarchical model formulation that yields excellent calibration results for multiple individual drivers, improving performance on a small data set over the multivariate normal model used by Mazinur. This capability is important because the use of one parameter set to represent the behavior and preferences of many drivers violates the assumptions of some models that parameters can only remain constant for a single driver \cite{treiber} or even a single CF instance \cite{newell}. Further, the ability to perform driver-specific calibrations is required for agent-based driver behavior modeling \cite{ear}. Second, we implement our ideas in TFP and Edward2, a software library and API for probabilistic programming that abstract away the complexity of MCMC algorithm implementation details, and that can easily scale to meet the compute demands associated with large data sets and simulator parameter search spaces. Finally, we discuss additional benefits of the application of Bayesian methods to car-following model calibration not addressed in previous work, including for example the utility of the Bayesian interpretation of p-values. 

\section{Background and Preliminaries}

\subsection{Bayesian Programming}

A Bayesian program (BP) is a generic formalism that can be used to describe many classes of probabilistic model, including Hidden Markov Models, Bayesian Networks and Markov Decision Processes \cite{diard}. This formalism organizes a graphical model, which encodes prior knowledge about the inference problem, together with model variables and observed data into a structure like the following:
\newline
\newline
\text{Program}
$\begin{cases}
\text{Desc.}
\begin{cases}
\text{Spec.}
\begin{cases}
\text{Variables} \\
\text{Decomposition} \\
\text{Forms (parametric or program)}
\end{cases}\\
\text{Identification (using data)}
\end{cases}\\
\text{Questions.}
\end{cases}$
\newline
\newline

The program must define 1) a means of computing the joint probability over its model, variables and data, and 2) a means of answering a specific inference question given that joint probability. For example, a Hidden Markov Model would have a sequence of states and observations as its variables, emission and transition matrices as its model, and various message-passing algorithms as its means of answering inference questions. The Baum-Welch algorithm \cite{Baum} would be its means of identification.

\subsection{Probabilistic Programming}

Probabilistic programming languages (PPLs) allow BPs to be implemented and have inference performed over their parameters using software. PPLs are like ordinary programming languages but they additionally allow variables' values to be randomly sampled from distributions, allowing the output of programs written in them to vary non-deterministically given the same input. They also allow variables' values to be conditioned on data (observations). In the BP formalism, the description of a probabilistic program is as follows: 
\newline
\newline
\text{Prog.}
$\begin{cases}
\text{Desc.}
\begin{cases}
\text{Spec.}
\begin{cases}
\text{Variables: $\theta$ \textit{(latent)} and \textbf{x} \textit{(obs)}} \\
\text{Decomp.: $P(\textbf{x}, \theta) \propto P(\theta) P(\textbf{x} | \theta)$} \\
\text{Form: \textit{Probabilistic Program}} \\
\end{cases}\\
\text{Identification: \textit{MCMC} or \textit{VI}}
\end{cases}\\
\text{Question: $P(\theta | \textbf{x}).$} \\
\end{cases}$
\newline
\newline

The decomposition states that the posterior joint probability of the model parameters and the observations is proportional to the product of the prior probability of the model parameters, $P(\theta)$, and the likelihood of the observations given the model parameters, $P(\textbf{x}|\theta)$. Because we treat each CF instance as independent, the likelihood can be further decomposed into products of the probabilities of each individual $x_i \in \textbf{x}$:
$$P(\textbf{x}|\theta) = \prod_{i=0}^{|\textbf{x}|} P(x_i|\theta).$$



The inference algorithms included with PPLs, such as Markov Chain Monte Carlo (MCMC) and variational inference (VI), can be run on arbitrary graphical models, as opposed to algorithms that run on a limited set of model classes for which they have been specially invented (e.g. Bayesian and Markov Networks). To construct a probabilistic variation of a deterministic mathematical model, such as a car-following model, one merely needs to implement the model as the likelihood function using the primitives of the PPL, with random variables (and their corresponding probability distributions) used to represent model parameters and the response variable, and ordinary mathematical operations used everywhere else. If the model can be implemented, then probabilistic parameter estimation can be performed in a plug-and-play fashion. The incredible flexibility of PPLs like Edward2 and PyMC3 \cite{pymc3} allow models of varying compositions to be implemented, from the piece-wise linear Newell '02 model \cite{newell} to the highly non-linear Wiedemann '99 (W99) model, which includes conditional function evaluation. We chose the IDM model to demonstrate our calibration approach due to its moderate simplicity and interpretability, plus its widespread use in practice \cite{hammit}.

\begin{figure}[!t]
  \centering
    \centering
      \includegraphics[scale=0.5]{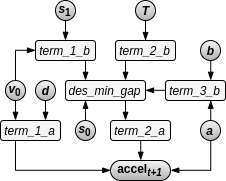}
  \caption{Directed acyclic graph representation of the probabilistic IDM model. Deterministic operations are abstracted into terms (rounded rectangles), and random variables (RVs) are represented by circles. The response variable, $accel_{t+1}$, is also treated as an RV because it is a function of RVs.}
  \label{fig:idm_dag}
\end{figure}

\subsection{The Intelligent Driver Model}
The Intelligent Driver Model \cite{treiber} predicts a following vehicle's next acceleration, $a_{t+1}$, given that vehicle's absolute velocity, $v$, and its velocity and distance relative to a leading vehicle, $\Delta v$ and $s$, at the current time:
\newline
\begin{equation}
    a_{t+1} = a\bigg[1 - \bigg(\frac{v}{v_0}\bigg)^{\delta} - \bigg(\frac{s^*(v,\Delta v)}{s}\bigg)^2\bigg],
\end{equation}
\newline
with:
\newline
\begin{equation}
    s^*(v,\Delta v) = s_0 + s_1 \sqrt{\frac{v}{v_0}} + \textit{T}v + \frac{v \Delta v}{2 \sqrt{a b}}.
\end{equation}
\newline
The estimable parameters are $v$, the desired velocity; $T$, the "safe" time headway, which is the time it would take the following vehicle to close the gap between itself and the leading vehicle; $a$, the maximum acceleration; $b$, the desired or "comfortable" deceleration, with higher values corresponding to more aggressive and late braking; $s_0$ and $s_1$, jam distance terms that correspond to stop-and-go traffic when values are low; and $\delta$, which represents the rate at which a driver will decrease acceleration as the desired velocity is approached.  This decrease in acceleration can range anywhere from exponential to linear or sub-linear.

IDM is a member of the class of collision-free car-following models. In ordinary situations, the vehicle will decelerate according to $b$, but in emergencies, deceleration can occur at an exponential rate \cite{ferreira}. Other widely used car-following models include the Gipps model \cite{gipps}, which is used in the AIMSUN simulator and is also a collision-free model, and Wiedemann '99, a psycho-physical model that is the basis of VISSIM \cite{barcelo}. IDM is itself included as an option in the deep reinforcement learning framework for traffic simulation, Flow \cite{flow}.

\subsection{Naturalistic Driving Data Set} \label{sec:naturalisticdrivingset}

Microsimulation models depend on observations of individual driver/vehicle characteristics such as velocity and inter-vehicle spacing, as opposed to macrosimulation models, which operate on aggregate measures like density and flow. The vehicles must be instrumented to collect data at this level of granularity with radar and other sensors, and their state (e.g., steering angle, acceleration, etc.) must be captured from their Controller Area Network (CAN) buses. This work was supported by data collected in such a fashion as part of a naturalistic driving study conducted by the Volpe Center. The original purpose of the study was to observe driver transits through construction zones, and use the observations to develop a work zone-specific car-following model. A subset of 207 quality-controlled car-following instances across 54 unique drivers of a single vehicle was selected from the total set of instances constructed from the raw data. Unused instances were plagued by temporal discontinuities resulting from hardware outages. The small data set size will help us demonstrate the utility of informative Bayesian priors.

\section{Calibration Method}

\subsection{Probabilistic Intelligent Driver Model Formulation}

A probabilistic variant of IDM can be easily derived in the context of the BP formalism, and conveniently constructed using a PPL. In this work, the latent variables are $\theta$ = \{$v_0$, \textit{T}, \textit{a}, \textit{b}, $\delta$, $s_0$, $s_1$\} and the observed variables are \textbf{x} = $\{s, v, \Delta v\} \cup \{a_{obs}\}$; our means of joint probability identification is a Hamiltonian Monte Carlo (HMC) variant of MCMC; and our form is the IDM model implemented as an Edward2 program.
\newline

We explore three probabilistic model formulations:
\begin{enumerate}
	\item a single multivariate-normal of dimension $k$ equal to the number of parameters, following the formulation in \cite{rahman}. The single set of parameters are calibrated using data from all drivers $d \in D$, where $|D|$ equals the number of drivers:
    \newline
    \newline
    $\theta_d \sim N_k(\mu,\Sigma)$.
	\newline
    \item a two-level hierarchy of univariate normals for which each driver has an individual set of $k$ random variables with their mean and standard deviation drawn from a pair of parent distributions shared across all drivers:
    \newline
    \newline
    $\theta_{d} = \mu_{d} + \sigma_{d} \: \times \: \theta_{d,norm}$;
    \newline
    \newline
    $\theta_{d,norm} \sim N_k^{L1}(0,1)$;
    \newline
    \newline
    $\mu_d \sim N_k^{L2}(\mu_\mu,\sigma_\mu)$;
    \newline
    \newline
    $\sigma_d \sim N_k^{L2}(\mu_\sigma,\sigma_\sigma)$.
    \newline
    \newline
    This arrangement allows each driver a unique parameterization that shares statistical strength with those of other drivers during calibration to avoid the consequences of insufficient data. We chose a non-centered implementation of the hierarchical model because it resulted in slightly lower error than a centered one that we explored but do not present here.
	\newline
    \item a set of $k$ univariate normals, each having an individual mean and standard deviation for each driver:
    \newline
    \newline
    $\theta_d \sim N(\mu_d,\sigma_d)$.
    \newline
    \newline
    This is the arrangement for which we expect model performance to suffer from a lack of training data (especially in cases where there may exist only one car-following instance for a driver.
    \newline
\end{enumerate} 
For simplicity, we assume there is no benefit to allowing the parameters of a single model to have different formulations.

\subsection{Modeling the Data Distribution}

For each model type, a normal distribution was chosen to represent the response variable, with each CF instance's distribution being assigned its corresponding observed mean and standard deviation. This assumption of normality is relatively safe given an inspection of each instance's quantile-quantile plot. While the plots do show heavy tails for several instances, in our experiments the use of normal data model performed better than Student's t-distribution models, indicating that the response distributions are not actually heavy-tailed but simply contain outliers. Each instance was as statistically independent, but with all time steps within a single instance being dependent.

\subsection{Calibration Procedure}
\subsubsection{Choosing an Inference Algorithm}
The inference algorithm chosen to identify the target joint probability distribution was Hamiltonian Monte Carlo, an advanced MCMC algorithm that uses gradients to inform its proposals for the next state to be explored during inference. HMC can be expected to require many fewer iterations to converge than alternative algorithms \cite{vehtari} This algorithm applies to the probabilistic IDM because the model equations are differentiable. A model such as W99 would require the use of an alternative algorithm like Metropolis-Hastings, also available in TFP, because of its conditionally invoked, driving regime-specific equations. 

\subsubsection{Testing for MCMC Convergence}
To determine at what number of burn-in steps that the search for the target joint distribution converges, the search was run multiple times with each successive run including an additional $1500$ steps, starting with $1500$ at the base run. This restart method was chosen after observing that, in TFP, initializing one search with the resulting state of a preceding search led to either slow or no progress being made. If the difference in joint probability between two consecutive runs fell below an arbitrary threshold, the results of the calibration of the preceding run were accepted as final.

\subsubsection{Scalability of Calibration Procedure}
Each probabilistic model calibration required no more than 9,000 iterations to converge, compared to reports of hundreds of thousands of iterations required in earlier literature for the same IDM model \cite{rahman}, even with only four out of seven of its parameters being calibrated. This speed (under four minutes) on a single quad-core CPU, coupled with the support for distributed computing across a virtually unbounded number of computers offered by TFP, make the application of our calibration method achievable for big data sets.

\begin{table}[ht]
\begin{center}
\caption{Measures of Calibration Error}
\label{tab:cnn_architecture_performance_comparison}
%
%
\begin{tabular}{p{1.4cm}p{.7cm}p{.7cm}p{.7cm}p{.7cm}p{.7cm}p{.7cm}}
\hline\noalign{\smallskip}
& \multicolumn{3}{l}{Root Mean Square Error} & \multicolumn{3}{l}{Average KL-Divergence}\\
\hline\noalign{\smallskip}
Prior $\sigma$ & 1 & 10 & 100 & 1 & 10 & 100 \\
\hline\noalign{\smallskip}
Pooled & .1842 & .1592 & .1529 & 397.88 & 213.16 & 194.67 \\
Hierarchical & \textbf{.1527} & \textbf{.1500} & \textbf{.1493} & \textbf{190.72} & \textbf{182.49} & \textbf{175.13} \\
Individual & .1864 & .1587 & .1548 & 492.59 & 212.63 & 194.71 \\
Literature\cite{treiber}* & \multicolumn{3}{c}{149.9} & \multicolumn{3}{c}{5.86e9} \\
\hline\noalign{\smallskip}
\multicolumn{7}{c}{$^*$ $\theta$ = \{$v_0$, \textit{T}, \textit{a}, \textit{b}, $\delta$, $s_0$, $s_1$\} = \{6.5, 1.6, .73, 1.67, 4., 2., 0.\}}
\end{tabular}
\end{center}
\end{table}

\begin{figure*}[t]  
\begin{center}  
\begin{tabular}{cc}  
\includegraphics[width=6.8in,]{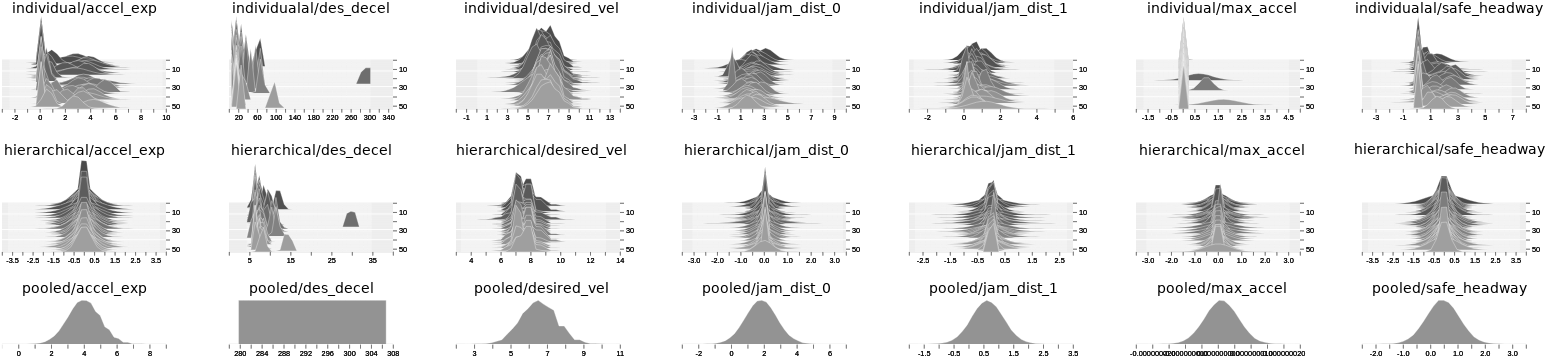}
\end{tabular}  
\caption{\label{fig:parameter_histograms} Posterior Model Parameter Distributions: \textit{Individual} parameters are presented in the first row, \textit{Hierarchical} in the second, and \textit{Pooled} in the third. In rows one and two, each of the 54 drivers' unique distributions are grouped and arrayed by parameter. The effect of constraining individual means and standard deviations to be sampled from a shared global mean and standard deviation is apparent when comparing the first and second rows. Note the differences in x-axis scale when measuring the visualizations, especially in the third row.}
\end{center}  
\end{figure*}

\section{Calibration Results}

\subsection{Error Analysis}

Four trends are apparent in the calibration results. First, the hierarchical model outperforms each of the other two models for every combination of prior $\sigma$ and measure of error. Second, the improvement of the hierarchical over the other models decreases monotonically with the decrease in the influence of the priors. Thirdly, as the prior $\sigma$ values increase, making them less informative to the search for the target distribution, the MoEs of all three models decrease significantly. Given that priors are known to influence posterior inferences less as data sets become larger and more representative of the phenomena they document \cite{gelman}, we can interpret this correlation between choice of $\sigma$ and fitness of model as indicating that the data set under consideration in this work is not sufficient in size to trust the under-regularized posterior distributions for use at test time, in spite of having a closer fit to the data. The ultimate test of performance remains direct observation of the simulated driving behavior resulting from the application of each model, a task reserved for future work. Finally, the prior influence trend in the hierarchical model is relatively weak compared to the other two models. The variant with the strongest regularization is only marginally less similar to the true response distribution than the variants with weaker regularization, making that first variant more attractive at test time.

\subsection{Analysis of Posterior Parameter Distributions}

Differences in parameter distributions resulting from the three modeling formulations can be seen in Figure \ref{fig:parameter_histograms}. We consider only those target joint distributions discovered using a prior $\sigma$ of 1. While distribution means vary quite dramatically across all \textit{Individual} parameters, \textit{Hierarchical} parameters are approximately zero-centered in all but two cases. Reasonably, the means of the \textit{Pooled} parameters appear to be near averages of the corresponding \textit{Individual} parameters, with each deviating from zero. For desired deceleration, a large variation across drivers is preserved among the \textit{Hierarchical} distributions, but they have been drawn an order of magnitude closer to zero than the \textit{Individual} distributions.

\section{Comparison of Bayesian and Genetic Algorithm Methods using Parameter Estimates}

The optimization technique pitted against Bayesian calibration in \cite{rahman} is not considered to be the industry standard technique \cite{hammit}. To further strengthen the case for our method, we compare it to our experience applying the industry standard method based on genetic algorithms. In particular, we use the implementation of a differential evolutionary algorithm available in TFP, with modifications made to allow the search space to be constrained to limit explosions in model parameter values. Rather than invest the necessary time in performing one instance of cross-validation for each iteration of algorithm hyperparameter tuning, we will attempt to compare calibration procedures without analytically comparing the models that they produce. We instead heuristically compare the estimates of model parameters that they produce. 

\subsection{The Differential Evolution Algorithm}

The differential evolution (DE) algorithm \cite{Kara} as implemented in TensorFlow Probability is initialized using a set of candidate solution parameter vectors called a population (as opposed to the single vector specification of the prior in the Bayesian method). An objective function is used to measure each candidate's fitness with respect to the model and data. The search through the parameter space proceeds by iteratively evaluating the population, and then performing two genetic operations on candidates: mutation (also known as crossover) and recombination. These perturbations encourage improvement in candidate fitness and exploration of the space, respectively. The algorithm depends on two hyperparameters: a differential weight that controls the magnitude of the mutation operation, and a crossover probability. Our fitness function of choice was the average of the root mean squared errors per car-following instance, plus an additional regularization term of the Euclidean distance between the candidate parameter values and the literature values specified in Table  \ref{tab:cnn_architecture_performance_comparison}. We introduce a third hyperparameter, lambda, which scales the regularization term. To discover the best solution achievable using the differential evolution algorithm, the best possible set of hyperparameter values must be identified with respect to the model and data. We used two search methods, grid search and Bayesian Optimization.

\subsection{Hyperparameter Tuning via Grid Search}

The first approach used to identify hyperparamaters that could yield excellent model parameter estimates was that of grid search. With grid search, we exhaustively test a fixed set of combinations of hyperparameter values, performing calibration once for each combination. We considered crossover probabilities between 0.1 and 0.9 with a step size of 0.2, differential weights between 0.1 and 1.9 with a step size of 0.2, and lambda between 0 and 0.0001 with step sizes of 0.0000025. The best combination found yielded an average RMSE of 0.118534155, outperforming the best Bayesian calibration result. But, we observed that for every oucome of the grid search, including the best one, at least one of the model parameters exploded in value, making the calibration results unconvincing with respect to the value ranges inferred from \cite{ferreira}.

\subsection{Hyperparameter Tuning via Bayesian Optimization}

Bayesian Optimization (BO) is one variant of sequential model-based optimization (SBMO), a class of algorithms that search a function space for functions with optimal outputs given some particular inputs. In each iteration of a search, SBMOs measure the utility of points in the parameter space and choose the point thought to yield the best output of a function in the function space evaluated on those parameters. We used an implementation of Bayesian Optimization written in Python and the scikit-learn machine learning library to search for near-optimal DE algorithm hyperparameters. Bayesian optimization can be considered to improve on grid search because it can require fewer function evaluations and because it can discover parameter values that fall inside the gaps in continuous space that woudl go unevaluated in a grid search.
We found that the BO method discovered hyperparameters with lower average population RMSE than grid search, with the lowest being 3.5764458 compared to grid search's 12.985725 average, but that the absolute lowest RMSE for an individual was higher than that of grid search: 0.33400983 versus 0.118534155.

\section{Discussion and Future Work}\label{futurework}

\subsection{Implementation of Models with Varying Complexity}

IDM is a moderately complex model, having 7 parameters and a relatively simple functional form. The flexibility of TFP can be nicely demonstrated by implementing a less complex model such as the piece-wise linear Newell '02, having just two parameters, and a more complex model such as W99 with its ten estimable parameters and four groups of conditionally-executable functions (one per driving regime). These models would require unique implementation of their calibration procedures relative to IDM. In the case of Newell, time is an estimable parameter and requires that at each iteration of inference a unique approximation of "truly observed" responses be created for comparison with predicted responses. In the case of W99, time steps must be processed sequentially rather than in parallel because the driving regime at each time step depends on the predicted acceleration response at the previous time step. Further, the non-trivially different models could help to demonstrate the breadth of the applicability of the Bayesian approach.

\subsection{Bayesian Model Validation and Comparison}

\subsubsection{Validation via Bayesian Information Criterion}

The standard best practice for validating statistical and machine learning models is to perform either \textit{K}-fold or leave-one-out (LOO) cross-validation (CV), especially when a data set is too small to expect a held-out test set to be equally representative of the data distribution as the training set. Cross-validation estimates a model's predictive accuracy on out-of-sample data by partitioning the total data set into \textit{K} subsets (with \textit{K} = 1 corresponding to LOO-CV),  performing \textit{K} distinct model fits with a disjoint subset held out each time, using that held-out data subset to measure the \textit{K}th model's performance, and finally averaging over the \textit{K} models' performance metrics. Among the probabilistic models presented in this paper, this method is useful for validating  the pooled model but not the hierarchical or individual models, since the latter two include sub-models with parameters estimated using only one data point. Fortunately, the Bayesian framework offers validation metrics called \textit{information criterion} that asymptotically approximate LOO-CV using the entire data set. Criterion of particular interest are 1) the Watanabe-Akaike information criterion (WAIC) \cite{Watanabe}, which operates on samples from the posterior log likelihood distribution one observation at a time, and then sums over all observations in the data set to yield a single measure, and 2) Pareto smoothed importance sampling-based LOO (PSIS-LOO), which improves on WAIC "in the finite case with weak priors or influential observations \cite{Vehtari2017}," which is precisely our scenario. 

\subsubsection{Validation via Bayesian Two-Sample Tests}

The traffic simulator calibration literature contains at least 25 examples of measures of error used to validate models. The closest to a consensus on the most generally applicable MoE is around the root mean squared error (RMSE). P-values have been used in the CFM calibration literature to compare parameter values estimated using data collected from different driving and environment conditions \cite{hammittdissertation}. In \cite{lloyd}, the authors 1) propose the use of maximum mean discrepancy two sample tests to validate probabilistic models in a way that exposes the regions where two compared distributions vary most in a visualizeable way, and 2) explain how p-values as interpreted from a Bayesian perspective more stringently criticise a model than under a frequentist interpretation. We interpret their description in the car-following context to mean that a frequentist p-value indicates how unlikely a set of data is to have been generated by a model given some parameterization, and that a Bayesian p-value indicates unlikeliness given the exact parameterization that resulted from a calibration. Further, the Bayesian p-values can be computed for distributions as well as point estimates (e.g. distribution means), making comparisons between the results of  Bayesian calibration and an alternative optimization-based calibration possible. We intend to explore the possibility that Bayesian two-sample tests may serve as a good generalized measure of error for car-following models.

\subsubsection{Model-Based Comparison of Bayesian- and Genetic Optimization-based Calibration Procedures}

While comparisons made in this paper between our proposed calibration procedure and a state of the art procedure were based on arbitrary judgement of how convincing the model parameter values resulting from those competing procedures were, a more sound approach would be to base comparisons on model validation metrics. Particularly, cross-validation could be performed on the pooled Bayesian model and on the genetic model, and then used to compare predictive accuracy between the models rather than to validate the calibration results of each model. Recall that use of the hierarchical model is prohibited by our inclusion of layer 1 groups that have only one data point.

\subsection{Application in a Traffic Simulator}

Users of traffic models and simulators commonly assume that all drivers have sufficiently similar behavior to perform a single parameter calibration and use the results to represent average driving behavior. Some documented attempts to deviate from this practice and instead perform one calibration per driver have led to the simulator producing excessive crashes and other unrealistic behavior \cite{ear}. One potential explanation for this failure is the dramatic reduction is data size when only a single driver is considered, leading to an over-fit model that does not generalize to the simulated driving scenarios of interest. In future work, we intend to show that our calibration procedure using a hierarchical statistical model not only fits the data well, while allowing for inter-driver variance, but also produces convincing simulated trajectories with help from the use of default parameter values as strong priors on the estimated values.  
\subsection{Can Simulations Derived from Ill-fit Models be Trusted?}

The utility of augmenting the randomness built into traffic simulators by incorporating the uncertainty in the data collection and calibration procedures, as well as a sub-model's approximation of reality, has been explored in the past \cite{molina}. Not considered in past work, however, is the fact that a probabilistic model fit on a data set will almost certainly not fit perfectly. This is usually desired in the interest of model generalization, but also means that combinations of parameter values drawn from the posterior joint distribution may yield unrealistic driving behavior. Intuitively, because the product of Bayesian calibration is a joint distribution over parameters and field data, if the marginal posterior distribution over the data matched the observed distribution perfectly, then the marginal distribution over the parameters would necessarily produce behavior no different than that observed in the data. If the fit is not perfect, can reasonable simulations be guaranteed? In future work, we intend to explore potential answers to the question: How many parameter samples must be drawn, fed into a simulator, and the resultant simulation traces validated before one can feel confident that \textit{any} sample drawn will yield valid results?

\subsection{Constructing a Large and Complete Data Set}

While the ability to incorporate prior domain knowledge into the calibration process is demonstrated in this work to be a strength, particularly when time, budget and human capital constraints result in a relatively small data set, analysis of Bayesian calibration based on a larger and more \textit{complete} data set is preferred. Completeness is described in \cite{hammittdissertation} to be the property of representing all driving situations to which any parameter in the model is relevant. We assume that our NDS data set is sufficiently complete based on the length and variety of routes and durations traveled by driver subjects during data collection, but could make a more convincing case for our method by performing a rigorous qualification of a data set. A larger data set would also make cross-validation (CV) a more comfortable exercise. In this paper, we forego the use of CV in order to avoid contributing additional uncertainty into the calibration results. Even when using the safest method of CV with respect to added uncertainty, leave-one-out (LOO), when a data set is very small its one-left-out subsets can yield inference results that deviate non-trivially from those of the full set \cite{vehtari}.

\section{Conclusion}\label{conclusion}

This paper presented a new method by which a car-following model may be calibrated using Bayesian inference and a hierarchical probabilistic model implemented in the probabilistic programming language Edward2. The method was framed using the Bayesian programming formalism. Calibration results using a microscopic naturalistic driving data set of 54 with varying numbers of CF instances and varying instance lengths were shown to improve when using a hierarchical formulation of the Intelligent Driver Model over individual and pooled formulations. Moreover, the utility of the incorporation of Bayesian priors as a form of regularization into the calibration process was demonstrated. Potential benefits of a Bayesian approach to CFM validation and comparison, and future directions of research, were also discussed. To enable convenient reproduction and criticism of our proposed method, and to encourage the exploration of probabilistic programming within the transportation research and engineering communities, the implementation source code has been made available on GitHub: \url{https://github.com/foabodo/pwie}.

\bibliographystyle{IEEEtran}
\bibliography{main.bib}

\begin{thebibliography}{10}
\providecommand{\url}[1]{#1}
\csname url@samestyle\endcsname
\providecommand{\newblock}{\relax}
\providecommand{\bibinfo}[2]{#2}
\providecommand{\BIBentrySTDinterwordspacing}{\spaceskip=0pt\relax}
\providecommand{\BIBentryALTinterwordstretchfactor}{4}
\providecommand{\BIBentryALTinterwordspacing}{\spaceskip=\fontdimen2\font plus
\BIBentryALTinterwordstretchfactor\fontdimen3\font minus
  \fontdimen4\font\relax}
\providecommand{\BIBforeignlanguage}[2]{{%
\expandafter\ifx\csname l@#1\endcsname\relax
\typeout{** WARNING: IEEEtran.bst: No hyphenation pattern has been}%
\typeout{** loaded for the language `#1'. Using the pattern for}%
\typeout{** the default language instead.}%
\else
\language=\csname l@#1\endcsname
\fi
#2}}
\providecommand{\BIBdecl}{\relax}
\BIBdecl

\bibitem{ear}
C.~D. Yang and T.~Morton, ``Trends of transportation simulation and modeling
  based on a selection of exploratory advanced research projects: Workshop
  summary report,'' Office of Operations Research and Development, Federal
  Highway Administration, U.S. Department of Transportation, Tech. Rep., Jul.
  2012.

\bibitem{hammit}
B.~Hammit, R.~James, and M.~Ahmed, ``A case for online traffic simulation:
  Systematic procedure to calibrate car-following models using vehicle data,''
  11 2018, pp. 3785--3790.

\bibitem{rahman}
M.~{Rahman}, M.~{Chowdhury}, T.~{Khan}, and P.~{Bhavsar}, ``Improving the
  efficacy of car-following models with a new stochastic parameter estimation
  and calibration method,'' \emph{IEEE Transactions on Intelligent
  Transportation Systems}, vol.~16, no.~5, pp. 2687--2699, Oct 2015.

\bibitem{higdon}
D.~Higdon, M.~Kennedy, J.~C. Cavendish, J.~A. Cafeo, and R.~D. Ryne,
  ``Combining field data and computer simulations for calibration and
  prediction,'' \emph{SIAM J. Sci. Comput.}

\bibitem{tran}
D.~{Tran}, M.~{Hoffman}, D.~{Moore}, C.~{Suter}, S.~{Vasudevan}, A.~{Radul},
  M.~{Johnson}, and R.~A. {Saurous}, ``{Simple, Distributed, and Accelerated
  Probabilistic Programming},'' \emph{arXiv e-prints}, p. arXiv:1811.02091, Nov
  2018.

\bibitem{bayarri}
M.~J.~Bayarri, J.~O.~Berger, G.~Molina, N.~Rouphail, and J.~Sacks, ``Assessing
  uncertainties in traffic simulation: A key component in model calibration and
  validation,'' \emph{Transportation Research Record}, vol. 1876, pp. 32--40,
  01 2004.

\bibitem{molina}
G.~Molina, M.~J. Bayarri, and J.~O. Berger, ``Statistical inverse analysis for
  a network microsimulator,'' \emph{Technometrics}, vol.~47, no.~4, pp.
  388--398, 2005.

\bibitem{zhong}
\BIBentryALTinterwordspacing
R.~Zhong, K.~Fu, A.~Sumalee, D.~Ngoduy, and W.~Lam, ``A cross-entropy method
  and probabilistic sensitivity analysis framework for calibrating microscopic
  traffic models,'' \emph{Transportation Research Part C: Emerging
  Technologies}, vol.~63, pp. 147 -- 169, 2016. [Online]. Available:
  \url{http://www.sciencedirect.com/science/article/pii/S0968090X15004222}
\BIBentrySTDinterwordspacing

\bibitem{treiber}
M.~Treiber, A.~Hennecke, and D.~Helbing, ``Congested traffic states in
  empirical observations and microscopic simulations,'' \emph{Physical Review
  E}, vol.~62, pp. 1805--1824, 02 2000.

\bibitem{newell}
\BIBentryALTinterwordspacing
G.~Newell, ``A simplified car-following theory: a lower order model,''
  \emph{Transportation Research Part B: Methodological}, vol.~36, no.~3, pp.
  195 -- 205, 2002. [Online]. Available:
  \url{http://www.sciencedirect.com/science/article/pii/S0191261500000448}
\BIBentrySTDinterwordspacing

\bibitem{diard}
J.~Diard, P.~Bessi{\`e}re, and E.~Mazer, ``A survey of probabilistic models
  using the bayesian programming methodology as a unifying framework,'' 2003.

\bibitem{Baum}
L.~E. Baum, T.~E. Petrie, G.~Soules, and N.~R. Weiss, ``A maximization
  technique occurring in the statistical analysis of probabilistic functions of
  markov chains,'' 1970.

\bibitem{pymc3}
J.~Salvatier, T.~V~Wiecki, and C.~Fonnesbeck, ``Probabilistic programming in
  python using pymc3,'' 01 2016.

\bibitem{ferreira}
B.~{Ferreira}, F.~A.~F. {Braz}, A.~A.~F. {Loureiro}, and S.~V.~A. {Campos}, ``A
  probabilistic model checking analysis of vehicular ad-hoc networks,'' in
  \emph{2015 IEEE 81st Vehicular Technology Conference (VTC Spring)}, May 2015,
  pp. 1--7.

\bibitem{gipps}
P.~Gipps, ``A behavioural car-following model for computer simulation,''
  \emph{Transportation Research Part B: Methodological}, vol.~15, pp. 105--111,
  04 1981.

\bibitem{barcelo}
J.~Barcelo, \emph{Fundamentals of Traffic Simulation}, 01 2010.

\bibitem{flow}
N.~Kheterpal, K.~Parvate, C.~Wu, A.~Kreidieh, Eug{\`e}ne, E.~Vinitsky, and
  A.~M. Bayen, ``Flow: Deep reinforcement learning for control in sumo,'' 2018.

\bibitem{vehtari}
\BIBentryALTinterwordspacing
A.~Vehtari, A.~Gelman, and J.~Gabry, ``Practical bayesian model evaluation
  using leave-one-out cross-validation and waic,'' \emph{Statistics and
  Computing}, vol.~27, no.~5, pp. 1413--1432, Sep 2017. [Online]. Available:
  \url{https://doi.org/10.1007/s11222-016-9696-4}
\BIBentrySTDinterwordspacing

\bibitem{gelman}
\BIBentryALTinterwordspacing
A.~Gelman, \emph{Prior Distribution}.\hskip 1em plus 0.5em minus 0.4em\relax
  American Cancer Society, 2006. [Online]. Available:
  \url{https://onlinelibrary.wiley.com/doi/abs/10.1002/9780470057339.vap039}
\BIBentrySTDinterwordspacing

\bibitem{Kara}
D.~Karaboǧa and S.~Okdem, ``A simple and global optimization algorithm for
  engineering problems: Differential evolution algorithm,'' \emph{Turkish
  Journal of Electrical Engineering and Computer Sciences}, vol.~12, pp.
  53--60, 01 2004.

\bibitem{Watanabe}
S.~Watanabe, ``Asymptotic equivalence of bayes cross validation and widely
  applicable information criterion in singular learning theory,'' \emph{J.
  Mach. Learn. Res.}

\bibitem{Vehtari2017}
\BIBentryALTinterwordspacing
A.~Vehtari, A.~Gelman, and J.~Gabry, ``Practical bayesian model evaluation
  using leave-one-out cross-validation and waic,'' \emph{Statistics and
  Computing}, vol.~27, no.~5, pp. 1413--1432, Sep 2017. [Online]. Available:
  \url{https://doi.org/10.1007/s11222-016-9696-4}
\BIBentrySTDinterwordspacing

\bibitem{hammittdissertation}
B.~Hammit, ``Methods to explore driving behavior heterogeneity using shrp2
  naturalistic driving study trajectory-level driving data,'' Ph.D.
  dissertation, University of Wyoming, 9 2018.

\bibitem{lloyd}
\BIBentryALTinterwordspacing
J.~R. Lloyd and Z.~Ghahramani, ``Statistical model criticism using kernel two
  sample tests,'' in \emph{Proceedings of the 28th International Conference on
  Neural Information Processing Systems - Volume 1}, ser. NIPS'15.\hskip 1em
  plus 0.5em minus 0.4em\relax Cambridge, MA, USA: MIT Press, 2015, pp.
  829--837. [Online]. Available:
  \url{http://dl.acm.org/citation.cfm?id=2969239.2969332}
\BIBentrySTDinterwordspacing

\end{thebibliography}

\end{document}